\pdfoutput=1
\def\year{2020}\relax
\documentclass[letterpaper]{article} 
\usepackage{aaai20}  
\usepackage{times}  
\usepackage{helvet} 
\usepackage{courier}  
\usepackage[hyphens]{url}  
\usepackage{graphicx} 
\newtheorem{myAssume}{Assumption}
\usepackage{amsmath}
\usepackage[ruled]{algorithm2e}
\usepackage{subfigure}
\usepackage{makecell}
\usepackage{multirow}

\title{Neural Inheritance Relation Guided One-Shot Layer Assignment Search}
\author{Rang Meng,\textsuperscript{\rm 1}\thanks{Equal contributions.This work is done when Rang Meng is an intern at Hikvision Research Institute.}
        Weijie Chen,\textsuperscript{\rm 2}\footnotemark[1]
        Di Xie,\textsuperscript{\rm 2} 
        Yuan Zhang,\textsuperscript{\rm 2} 
        Shiliang Pu\textsuperscript{\rm 2}\thanks{Corresponding Author}\\ 
\textsuperscript{\rm 1}College of Control Science and Engineering, Zhejiang University\\
\textsuperscript{\rm 2}Hikvision Research Institute\\
r\_meng@zju.edu.cn,  \{chenweijie5, xiedi, zhangyuan, pushiliang\}@hikvision.com 
}

\begin{document}

\maketitle
\section{Abstract}
Layer assignment is seldom picked out as an independent research topic in neural architecture search. In this paper, for the first time, we systematically investigate the impact of different layer assignments to the network performance by building an architecture dataset of layer assignment on CIFAR-100. Through analyzing this dataset, we discover a neural inheritance relation among the networks with different layer assignments, that is, the optimal layer assignments for deeper networks always inherit from those for shallow networks. Inspired by this neural inheritance relation, we propose an efficient one-shot layer assignment search approach via inherited sampling. Specifically, the optimal layer assignment searched in the shallow network can be provided as a strong sampling priori to train and search the deeper ones in supernet, which extremely reduces the network search space. Comprehensive experiments carried out on CIFAR-100 illustrate the efficiency of our proposed method. Our search results are strongly consistent with the optimal ones directly selected from the architecture dataset. To further confirm the generalization of our proposed method, we also conduct experiments on Tiny-ImageNet and ImageNet. Our searched results are remarkably superior to the handcrafted ones under the unchanged computational budgets. The neural inheritance relation discovered in this paper can provide insights to the universal neural architecture search.

\begin{figure}[!h]
\centering
\subfigure[]{
\label{fig:subfig:a} 
\includegraphics[width=.9\columnwidth]{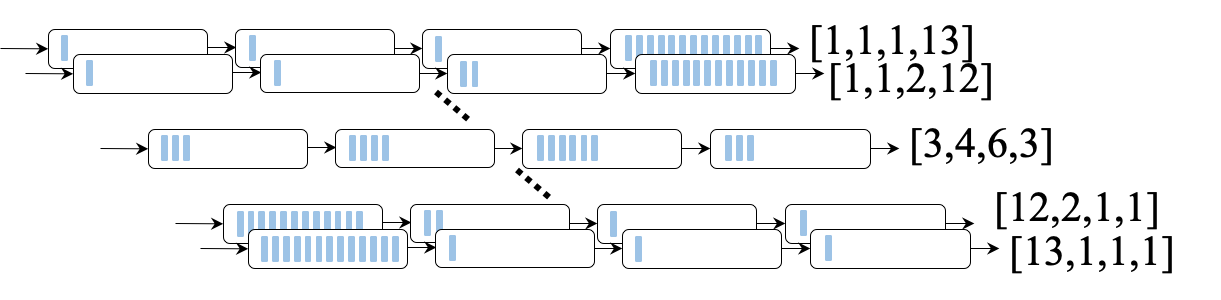}}
\subfigure[]{
\label{fig:subfig:b} 
\includegraphics[width=.9\columnwidth]{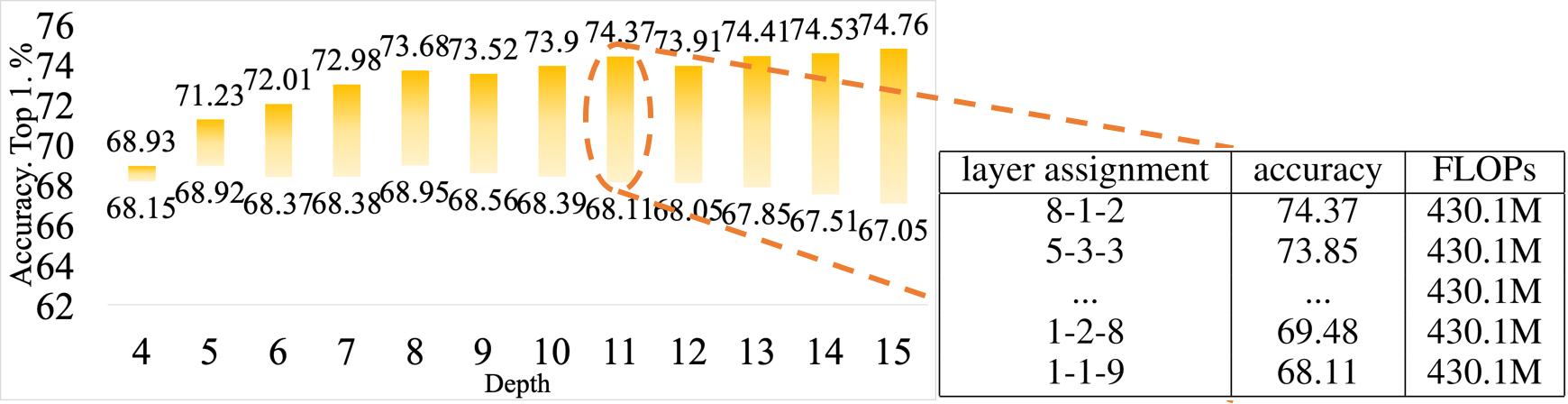}}
\subfigure[]{
\label{fig:subfig:c} 
\includegraphics[width=.9\columnwidth]{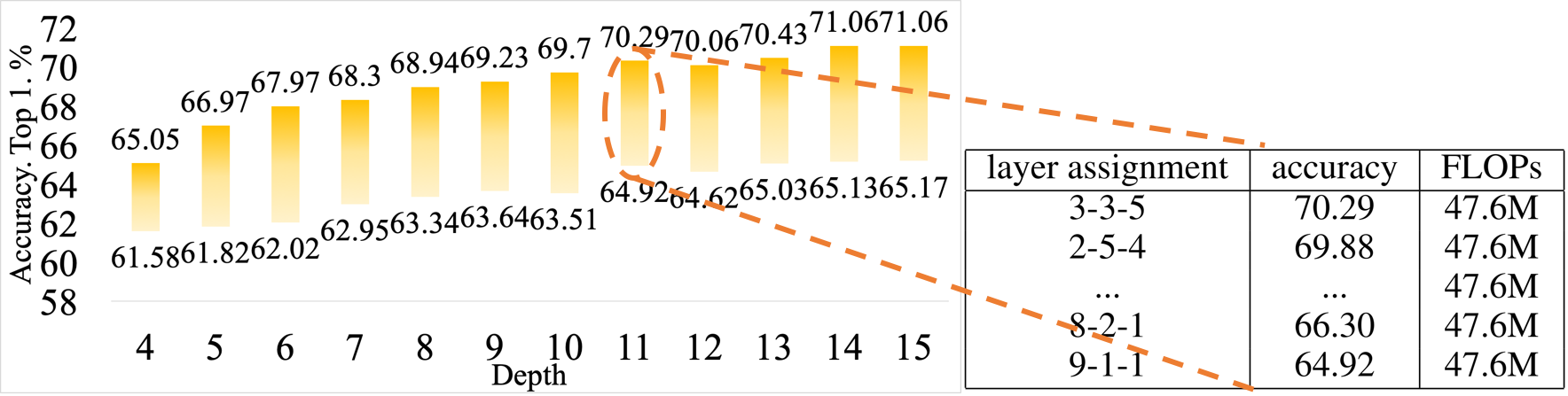}}
\caption{Different layer assignments, different network performances. (a) presents different layer assignments. All the networks follow the rule of \emph{half the spatial size, double the channels} to keep them with the same FLOPs. (b) and (c) illustrate the performance gap (the bin in these figures) among different layer assignments of plain and residual networks on CIFAR-100, respectively.}
\label{Figure 1} 
\end{figure}

\section{Introduction}
Network depth is a critical dimension in neural architecture design. An increasing network depth in convolutional neural networks (CNNs) will bring a remarkable performance boost, such as the accuracy increase from ResNet-18 to ResNet-152. However, an important problem is neglected: \emph{Given a fixed network depth, how to assign the layers into different groups for computing feature maps with different spatial resolution?} Taking ResNet-50\cite{he2016deep} as an example, it assigns [3, 4, 6, 3] stacked residual bottlenecks to compute feature maps with spatial resolution $56\times56$, $28\times28$, $14\times14$ and $7\times7$, respectively. In this paper, we name this process as layer assignment for short. Obviously, this is an empirically handcrafted design and cannot guarantee the best performance. As shown in Fig.\ref{Figure 1}, there exist many layer assignment manners with the same network depth whose accuracy in a given task are different a lot. Actually, the optimal layer assignment is changeable among different tasks and it is costly to search the optimal layer assignment for each task in a trial-and-error way. Therefore, it is essential to search the optimal layer assignment automatically and efficiently.

Before designing the method to search the optimal layer assignment, we are curious about the relation among different layer assignment. Is there any inheritance relation among the optimal layer assignments from shallow networks to deeper networks? Bringing this question, we build a neural architecture dataset of different layer assignments, which consists of 908 different neural networks trained on CIFAR-100, including plain networks and residual networks(we will release later). We separately pick out the optimal ones for each network depth from this dataset. Surprisingly but reasonably, as is shown in Fig.\ref{Figure 4}, we find that the optimal layer assignments for deeper networks are always developed from those for shallow ones. We name this as neural inheritance relation in this paper.

Inspired by the discovered neural inheritance relation, we propose a one-shot layer assignment search method via inherited sampling. First of all, one-shot method\cite{bender2018understanding} is an efficient neural architecture search framework, which constructs a supernet coupled every candidate sub-network in a weight-sharing manner. It can be trained through a sampling way (like uniform sampling\cite{guo2019single}). After that, the performances of the sub-networks sampled from this supernet are ranked so as to search the best one. In this paper, we couple the sub-networks with different layer assignments into a shared supernet, which can efficiently reduce training cost. We propose a neural inheritance relation guided sampling method to train the supernet. Starting from training shallow sub-networks sampled from the supernet, we search the optimal layer assignment and take it as a strong sampling priori to sample and train the deeper one which is named as inherited sampling for short. Different from other one-shot methods, we iterate the training and searching process from shallow to deep networks. In this way, our method can reduce the search complexity from $\mathcal{O}(m^n)$ to $\mathcal{O}(mn)$, where $m$ means the layer number and $n$ denotes the group number.

We conduct extensive experiments on CIFAR-100 to demonstrate the efficiency and accuracy of our proposed layer assignment search method. Our searched layer assignment for different network depth is strongly consistent with the optimal one selected from the architecture dataset. To the best of our knowledge, we are the first one to use architecture dataset to evaluate the performance of searching method. Before our work, other related works evaluate their performance through comparing with each other, and they never know the accuracy gap between their searched models and the best model actually. Moreover, to make sure the generalization of our proposed method, we also conduct extensive experiments on Tiny-ImageNet and ImageNet. The performance of our searched models surpass the corresponding handcrafted ones by a large margin. 

To summarize, our contributions can be listed as follows:

\begin{itemize}
\item We build an architecture dataset of layer assignment to analyze their hidden relation, and we discover a neural inheritance relation among the optimal layer assignment for each network depth. To the best of our knowledge, we are the first one to systematically investigate the impact of layer assignment to network performance.
\item We propose a neural inheritance relation guided one-shot method for automatic layer assignment search.
\item To the best of our knowledge, we are the first one to evaluate the performance of searching method through architecture dataset by providing the ground-true performance comparison of different networks.
\item Our searched layer assignments surpass handcrafted ones by a large margin.
\end{itemize}

\section{Related Works}
\subsection{Layer Growing}
The original training method of VGGNet \cite{simonyan2014very} may be the prototype of layer growing, which trains the deeper network inheriting the trained weights from the shallow one and only initializes the newly-stacked layers. Network Morphism \cite{wei2016network,Chen2016Net2Net,wei2017modularized} morphs a shallow network to a deeper network through stacking new layers initialized as identity mapping so as to expand its learning capacity. It can fast learn extra knowledges directly based on the trained shallow network. Beyond, AutoGrow \cite{wen2019autogrow} proposes to automatically grow the layers until the network accuracy stops increasing. Also, it proves that a simple random initialization is better than Network Morphism. 

The abovementioned methods only focus on how to initialize the newly-stacked layers and how to discover the network depth when the accuracy stops increasing. Different from them, our proposed method aims to search the optimal layer assignment to achieve the best network accuracy with unchanged computational budgets (network depth and FLOPs). Also, our method can search the optimal layer assignment from shallow to deeper networks once.

\subsection{Neural Architecture Search}
Neural Architecture Search (NAS) can search the network architecture automatically, including network depth, through reinforcement learning \cite{zoph2016neural,cai2018path,zoph2018learning,tan2019efficientnet,tan2019mnasnet}, evolution methods \cite{liu2017hierarchical,real2019regularized,perez2019mfas,elsken2018efficient} and gradient-based methods \cite{shin2018differentiable,luo2018neural,liu2019auto,wu2019fbnet,xie2018snas}. Layer Assignment Search (LAS) is a sub-problem of NAS. Before our work, no one purely picks out this sub-problem and analyzes its influence to the network performance. Different from other related NAS works, we build an architecture dataset about layer assignment to systematically analyze the hidden relation among different architectures, which inspires our proposed solution to LAS.

The most closely related NAS methods to ours is the one-shot methods \cite{bender2018understanding,brock2017smash,guo2019single,chu2019fairnas,yu2019network}, which couple each candidate network into a supernet in a weight-sharing manner. The two most important operations are a sampling method to train the supernet and an efficient method to search the best one. For the former one, there exist sandwich rule \cite{yu2019network}, uniform sampling \cite{guo2019single}, fairness-aware sampling methods \cite{chu2019fairnas} and so on. For the latter one, there exist greedy\cite{yu2019network}, evolutionary\cite{guo2019single}, differentiable searching methods\cite{stamoulis2019single}. These two operations are applied to make the candidate networks predictable and search the best one efficiently. However, since there are too many networks coupled into a shared supernet, it is too hard to make sure the candidate networks predictable. Also, the candidate network space is too large to search efficiently. 

How to reduce the candidate network space in training and searching phase is a feasible direction to solve these two problems. Different from the related methods, our proposed inherited sampling method to train the supernet can reduce the network space by a large margin.

\section{Problem Definition}

In the convolutional neural networks, the feature maps are usually downsampled into several spatial resolutions by strided convolution or pooling operations from shallow to higher layers. From this perspective, a CNN architecture can be divided into several groups. The intermediate feature maps in each group share the same spatial resolution. Let $a_{i}$ be the number of layers in the $i^{th}$ group. A CNN $\mathcal{A}^{m}_{n}$ with $m$ layers and $n$ groups ($m\geq n$) can be expressed as:
\begin{equation}
\begin{aligned}
&\mathcal{A}^{m}_{n}=[a_{1},a_{2},...,a_{n}] \\
&s.t.\quad
\begin{cases}
\sum^{n}_{i=1}a_{i}=m\\
a_i>0, i=1,2,...,n
\end{cases}
\end{aligned}
\label{definition}
\end{equation}

Apparently, there exist many solutions in Eqn.\ref{definition} especially as $m$ increases given a fixed $n$. Different solutions correspond to different layer assignment to the CNN architecture, and further lead to different performances in each task. In this work, a layer assignment search problem can be defined as to find an optimal solution of $\mathcal{A}^{m}_{n}$ with the best performance in each task, which can be formulated as:
\begin{equation}
\tilde{\mathcal{A}^{m}_{n}}=\mathop{\arg\max}_{\mathcal{A}^{m}_{n}}ACC_{val}(\mathcal{N}(w, {\mathcal{A}}^{m}_{n}))
\label{optimization}
\end{equation}
where $\mathcal{N}(w, {\mathcal{A}}^{m}_{n})$ means a network with trained weights $w$ and layer assignment ${\mathcal{A}}^{m}_{n}$, and $ACC_{val}(\cdot)$ denotes the accuracy on the validation dataset.

Theoretically, a naive enumeration method can solve this problem, in which networks with every enumerated layer assignment are trained and evaluated. However, it is too expensive to train every possible network. The number of the satisfied networks can be determined as:
\begin{equation}
\mathcal{O}=\frac{(m-1)!}{(n-1)!(m-n)!}
\label{complexity}
\end{equation}
where $!$ denotes factorial. In practice, it is usually necessary to search the best layer assignment in a range of network depth instead of a given $m$. Hence, the more generalized layer assignment search problem should be formulated as:
\begin{equation}
\{\tilde{\mathcal{A}}^{i}_{n}\}=\mathop{\arg\max}_{\mathcal{A}^{i}_{n}}ACC_{val}(\mathcal{N}(w, {\mathcal{A}}^{i}_{n})), i\in [n, m]
\label{optimization2}
\end{equation}

The complexity of enumeration method increases to:
\begin{equation}
\mathcal{O}=\sum_{i=n}^{m}\frac{(i-1)!}{(n-1)!(i-n)!}=\frac{\sum_{i=n}^{m}\prod_{j=1}^{n-1}i-j}{(n-1)!}
\label{complexity2}
\end{equation}
which can be represented as a complexity $\mathcal{O}(m^n)$ for short. Our work here exactly focuses on reducing the complexity and searching the optimal layer assignment efficiently.

In order to purely discuss the function of different layer assignment manners, we double the output feature maps when downsampling the spatial resolution in half to guarantee the invariant FLOPs among different layer assignments with the same network depth. 

\begin{figure}[!t]
\centering
\includegraphics[width=.9\columnwidth]{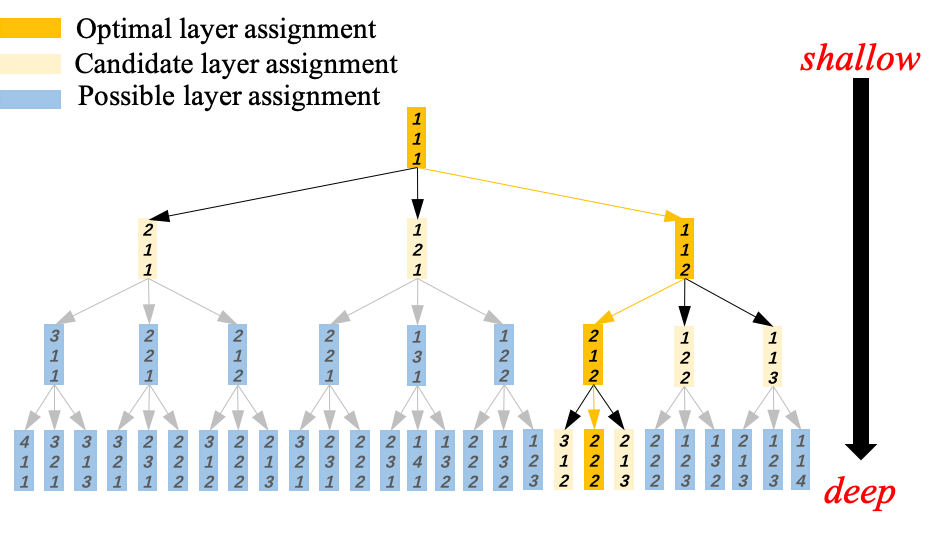} 
\caption{An intuitive visualization of network space reduction brought by neural inheritance relation.}
\label{Figure 2}
\end{figure}

\section{Method}
\subsection{Neural Inheritance Relation Assumption}

\begin{figure*}[!t]
\centering
\includegraphics[width=.75\textwidth]{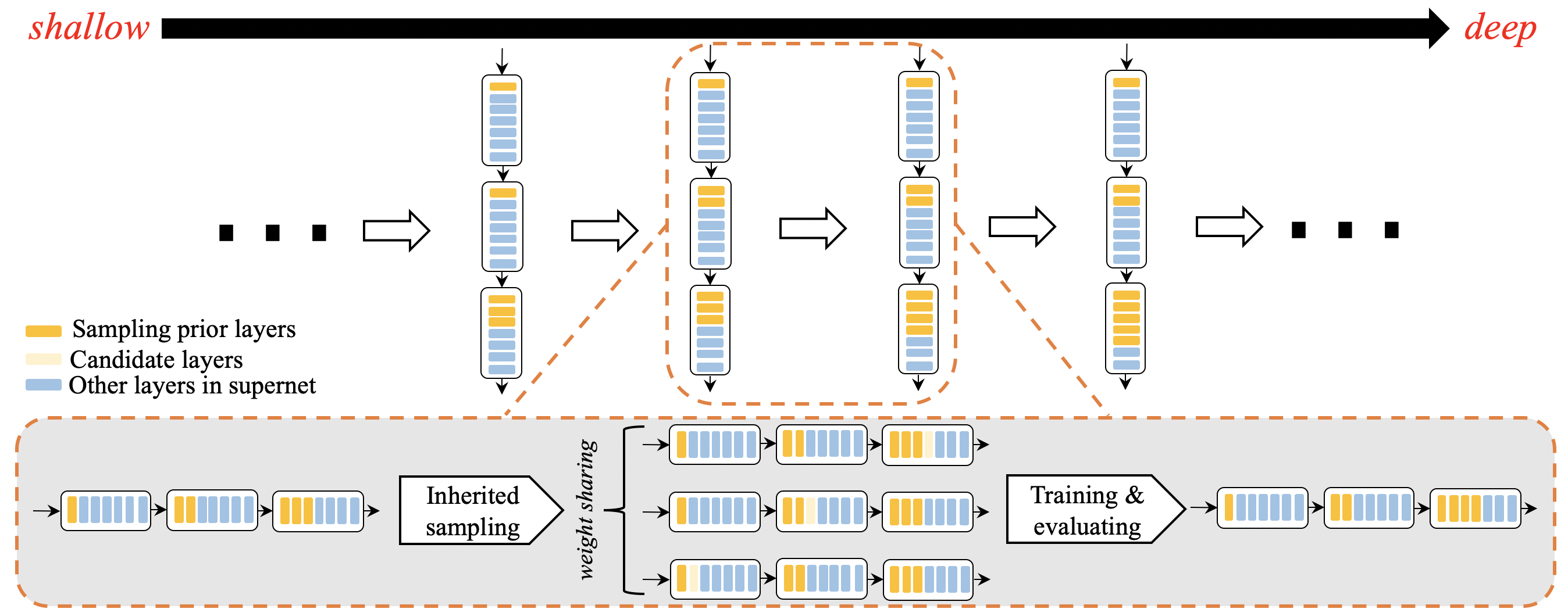} 
\caption{The pipeline of our proposed neural inheritance relation guided one-shot layer assignment search method.}
\label{Figure 3}
\end{figure*}

As shown in Eqn.\ref{complexity2}, the layer assignment space for searching is very enormous. Here naturally comes a question: \emph{does there exist redundancy in this layer assignment searching space?} We believe there must exist some hidden relation among different layer assignments. Given the performance comparison of a small portion of them, it can directly derive the performance comparison of other layer assignments without training, which further reduces the searching space. To proceed, we propose a reasonable Neural Inheritance Relation Assumption (NIR assumption).

{\myAssume{The optimal layer assignments for deep networks always inherit from those for shallow networks.}\label{assume}}

Mathematically, this assumption can be formulated as:
\begin{equation}
\!\tilde{A}^{m}_{n}\!\!=\!\!\tilde{A}^{m-1}_{n}\!+\!\mathop{\arg\max}_{\textbf{1}}ACC_{val}(\mathcal{N}(w, \tilde{A}^{m-1}_{n}\!+\!\textbf{1})), \textbf{1}\!\!\in\!\!\Omega\!\!
\end{equation}
where $\textbf{1}$ is a $n$-dimension one-hot vector and $\Omega$ is collection of all one-hot vectors. This formula means we can search the optimal layer assignment of $m$ layers based on the result of $m-1$ layers. According to this assumption, we can search the optimal layer assignment from shallow to deep networks in a recursive process, which reduces the searching complexity from $\mathcal{O}(m^n)$ to $\mathcal{O}(mn)$ as shown in Fig.\ref{Figure 2}.

In the experimental part, we will prove it by building an architecture dataset of different layer assignments.

\subsection{Neural Inheritance Relation Guided One-Shot LAS Method}

Based on the NIR assumption, the layer assignment can be searched recursively from shallow to deep networks. To accelerate the search process, a weight-sharing mechanism is integrated, where a supernet (one-shot model) is constructed covering every possible layer assignment. That is, a given layer assignment can be sampled from this supernet without training stand alone to reduce the training time.


Moreover, according to the NIR assumption, it is unnecessary to couple every possible network into the supernet causing a waste of training resource. Actually, the more sub-networks coupled into the supernet, the larger the accuracy gap between one-shot and stand-alone models will be. We define the accuracy gap as:
\begin{equation}
\begin{aligned}
GAP=\frac{1}{|\Gamma|}\sum_{A\in \Gamma}^{}|ACC_{val}(\mathcal{N}(w_{stand-alone}, A))\\
-ACC_{val}(\mathcal{N}(w_{one-shot}, A)| 
\end{aligned}
\label{accuracy_gap}
\end{equation}
where $\Gamma$ is the collection of layer assignments and $|\Gamma|$ denotes the corresponding count. $w_{stand-alone}$ and $w_{one-shot}$ denote the weights for stand-alone training and supernet training. We claim that only if the accuracy gap is small enough will the performance comparison among sub-networks be highly confident. However, the more coupled sub-networks, the larger this accuracy gap will be. Here we propose to train the supernet through a Neural Inheritance Relation Guided Sampling method, where only a few sub-networks couple into the supernet for each network depth.

This sampling method is inspired by NIR assumption. Given the searched optimal layer assignment $\tilde{A}^{m-1}_{n}$ for the network with $m-1$ layers, there exist $n$ inherited layer assignment candidates for the network with $m$ layers, namely $\{\tilde{A}^{m-1}_{n}+\textbf{1}\}$. We only sample $n$ candidate networks from supernet to train uniformly. After several epochs training, we evaluate the performance of these candidate networks on the validation dataset and then select the best one as the optimal layer assignment $\tilde{A}^{m}_{n}$ for the network with $m$ layers. Repeatedly, $\tilde{A}^{m}_{n}$ can be taken as a strong sampling priori to search $\tilde{A}^{m+1}_{n}$ in next step. The entire procedure is shown as Fig.\ref{Figure 3}.

\subsection{Discussion}
Previous one-shot search methods can be divided into two steps: 1) Training the supernet coupled every possible sub-network; 2) After training, evaluating the sub-networks sampled from the supernet and then searching the best one. 

Compared with these kinds of methods, our proposed method turns these two steps into an integration, which actually cuts an enormous sub-network space into many small pieces and then picks the useful ones from shallow networks to deeper networks for training and searching in a recursive way. As is mentioned above, our proposed method extremely reduces the searching complexity from $\mathcal{O}(m^n)$ to $\mathcal{O}(mn)$, which makes the one-shot training faster and brings more confident searching results.

\section{Experiments}
In this work, we mainly conduct experiments on CIFAR-100 to analyze the neural inheritance relation for both plain networks and residual networks. Each layer in these networks follow the rule of \emph{half spatial resolution, double channel size} so as to keep FLOPs unchanged for different layer assignments with the same network depth. By building an architecture dataset of layer assignment on CIFAR-100, we empirically prove our proposed Assumption \ref{assume}. We use this architecture dataset to evaluate our proposed LAS method. To demonstrate the generalization of our proposed method, we also carry out experiments on Tiny-ImageNet and a very large-scale dataset ImageNet to further analyze: 1) The different impact of layer assignments on shallow networks and very deep ones. 2) The layer assignments of the same network searched for different datasets. 3) The improvement of searched layer assignments to handcrafted counterparts.

\begin{algorithm}[!t]
\caption{Layer Assignment Search Algorithm}
\LinesNumbered
\KwIn{
groups number $n$, target layers number and the deepest layers number in supernet $m_t$ and $m_s$, step size $K$, training data loader $D_t$, validation dataset $D_v$, 1000-images subset of training data loader $D_s$}
\KwOut{
the optimal layer assignments $\tilde{\mathcal{A}}^{m_t}_{n}$}
Warm up the deepest sub-network $\mathcal{N}(w, \mathcal{\dot{A}}^{m_s}_{n})$\;
Set the seed layer assignment$\tilde{\mathcal{A}}^{n}_{n}$\;
\For{$p=n$ \textbf{to} $m_t$}{
    Inherited sampling, $\{(\tilde{\mathcal{A}}^{p}_{n}+\textbf{1}_i),i=1,2,\ldots,n\}$\;
    \For {$d=1$ \textbf{to} $K$}{
        \For{$data$, $labels$ \textbf{in} $D_t$}{
            Train$(\mathcal{N}(w, \mathcal{\dot{A}}^{m_s}_{n}), data, label)$\;
            \For{$i=1$ \textbf{to} $n$}{
                Train($\mathcal{N}(w, \tilde{\mathcal{A}}^{p}_{n}+\textbf{1}_i), data, label)$\;
            }
        }
    }
    \For{$i=1$ \textbf{to} $n$}{
        Recalculate\_BN($\mathcal{N}(w, \tilde{\mathcal{A}}^{p}_{n}+\textbf{1}_i), D_s$)\;
        $ACC_i$=Predict($\mathcal{N}(w, \tilde{\mathcal{A}}^{p}_{n}+\textbf{1}_i), D_v$)\;
    }
    $\tilde{\mathcal{A}}^{p+1}_{n}$=Top1(\{($\tilde{\mathcal{A}}^{p}_{n}+\textbf{1}_i, ACC_i), i=1,2,\ldots,n$\})\;
}
\end{algorithm}

\subsection{Benchmark Datasets}
\subsubsection{CIFAR-100\cite{krizhevsky2009learning}}is a dataset for 100-classes image classification. There are 500 training images and 100 testing images per class with resolution $32\times32$. In this dataset, we build a serial of VGG-like plain networks and a serial of residual networks to analyze the neural inheritance relation. We divide the feature extractor of each network into 3 groups with 3 input resolutions $32\times32$, $16\times16$ and $8\times8$. For the plain networks, the channel of each group corresponds to 64, 128 and 256. Each group consists of several stacked $3\times3$ convolution layers and ends with a max-pooling layer for downsampling. A batch normalization and a ReLU layer are appended after each convolution layer. The classifier consists of 3 fully-connected layers with channel 512, 512 and 100. For the residual networks, we exactly follow the design of \cite{he2016deep} for CIFAR-10 with channels 16, 32, 64 for each group. Different from the plain networks, the basic unit of residual networks is bottleneck with two $3\times3$ convolution layers. LAS problem is to determine the number of layers (convolution layer, bottleneck or other basic units) assigned for each group given a target network depth.

\subsubsection{Tiny-ImageNet}is a subset of ImageNet for 200-classes image classification. There are 500 training images, 50 validation images and 50 testing images per class with resolution $64\times64$. In this dataset, we adopt two famous networks MobileNet-V1 and ResNet-50 as the representatives to show the efficiency of our LAS method, which covers the cases including shallow network, very deep network, plain network and residual network. The handcrafted layer assignments for MobileNet-V1 and ResNet-50 are [2, 2, 6, 2] and [3, 4, 6, 3] with resolution $16\times16$, $8\times8$, $4\times4$ and $2\times2$.

\subsubsection{ImageNet\cite{russakovsky2015imagenet}}is a 1000-classes image classification dataset, which consists of 1.28 million images for training and 50k for validation. In this dataset, we mainly adopt MobileNet-V1 as a representative to illustrate the generalization of our proposed LAS method to the very large-scale dataset since it is a great tendency to search a compact light-weight network with high performance.

\subsection{Implementation Details}
To simplify the description of our algorithm, we name the searching process from a network depth to next network depth as one \emph{step}. To balance the training iteration of each step, we warm up the deepest sub-network in supernet through training $60$ epochs by a pre-set base learning rate before searching. After that, we initialize the seed layer assignment as $[1]\times n$ for searching. During training the supernet in each step, we iteratively train the deepest sub-network and $n$ sub-networks obtained by inherited sampling. The reason why we always train the deepest sub-network is to tune the parameters uncovered by the sampled sub-networks in real time to accelerate the searching process for the following steps. In each step, we adopt half of the base learning rate with linear decaying policy for 10 epochs training. In the end of each step for performance evaluation, we re-calculate all the statistics in the batch normalization layer with 1000 images randomly selected from the training dataset. Note that we only preserve one optimal layer assignment as the inherited sampling priori for the next network depth. Our entire searching process is summarized in Algorithm 1.

\subsection{Experiments on CIFAR-100}
\subsubsection{Training Details.} During training phase, we first zero-pad the images with 4 pixels on each side and then randomly crop them to produce $32\times32$ images, followed by randomly horizontal flipping. We normalize them by channel means subtraction and standard deviations division for both training dataset and validation dataset. During building an architecture dataset of layer assignment, we train all the enumerated networks in Pytorch. using SGD with Nesterov momentum 0.9. The base learning rate is set to 0.1 and multiplied with a factor 0.2 at 60 epochs, 120 epochs and 160 epochs, respectively. Weight decay is set as 0.0005. All the networks are trained with batch size 128 for 200 epochs. During searching, the setup is the same as above except that the base learning rate is set to 0.05 with linear decaying policy for each step as shown in Fig.\ref{Figure 6}.

\subsubsection{Architecture Dataset Construction.} 
\begin{figure}[!t]
\centering
\subfigure[]{
\label{fig:subfig:a}
\includegraphics[width=.9\columnwidth]{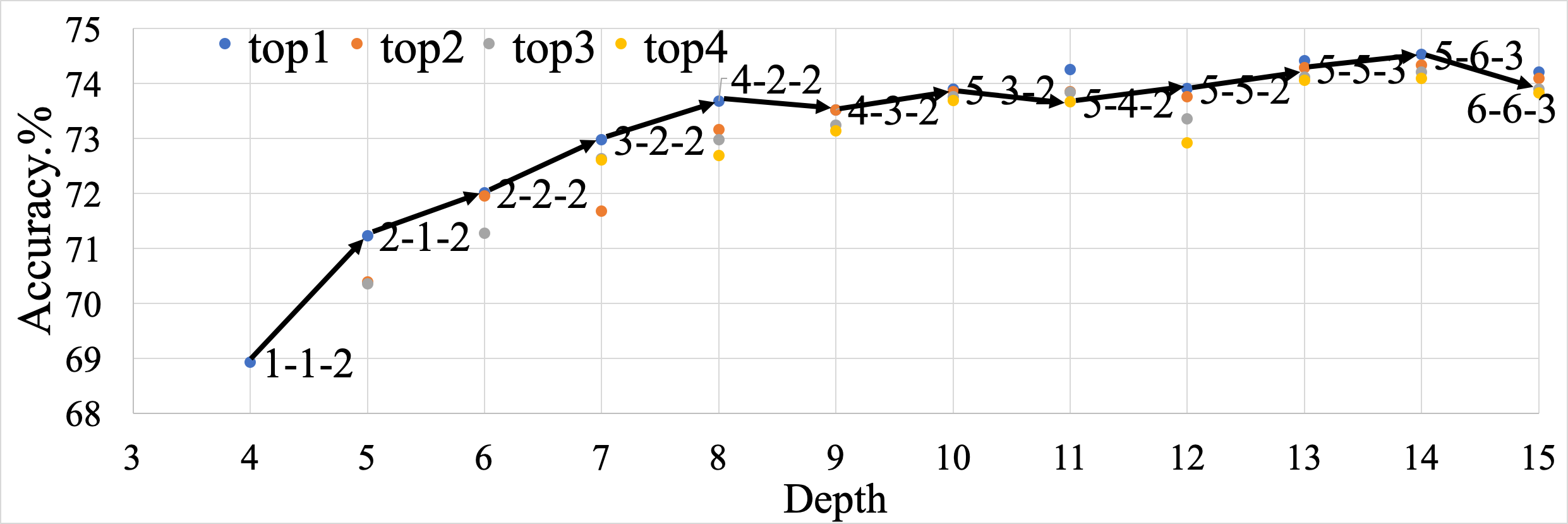}}
\subfigure[]{
\label{fig:subfig:b}
\includegraphics[width=.9\columnwidth]{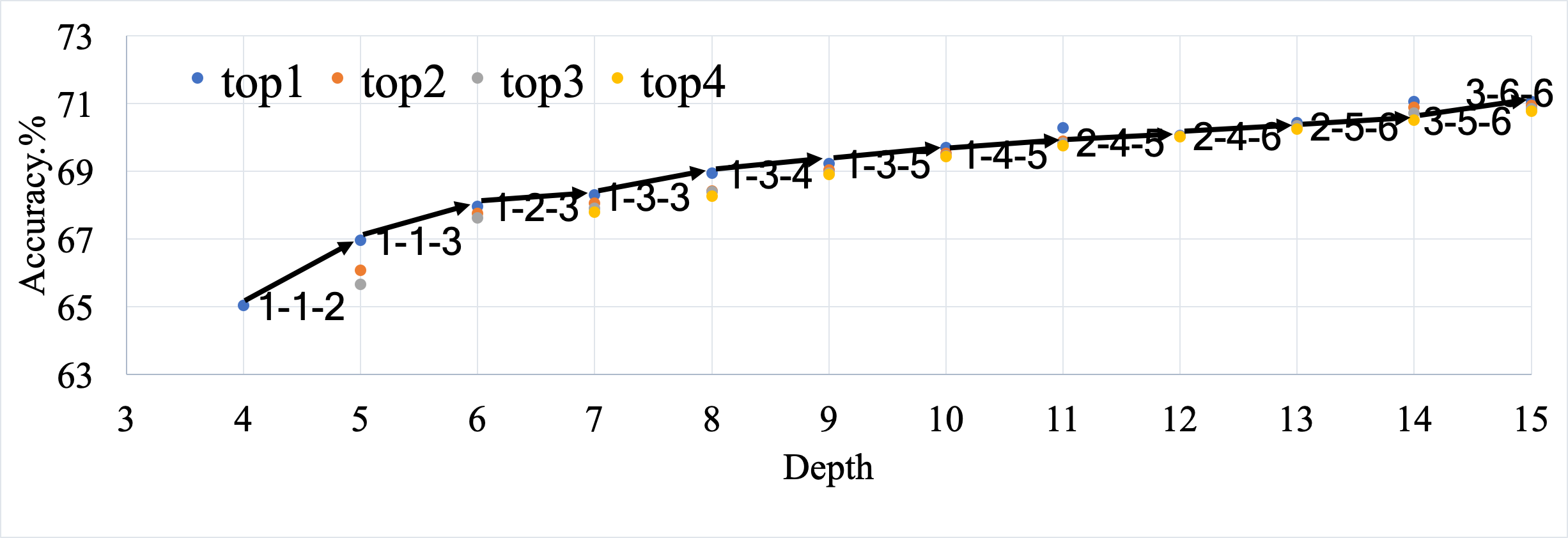}}
\caption{Neural inheritance relation in plain (a) and residual (b) networks on CIFAR-100.}
\label{Figure 4}
\end{figure}

\begin{figure}[!t]
\centering
\includegraphics[width=.9\columnwidth]{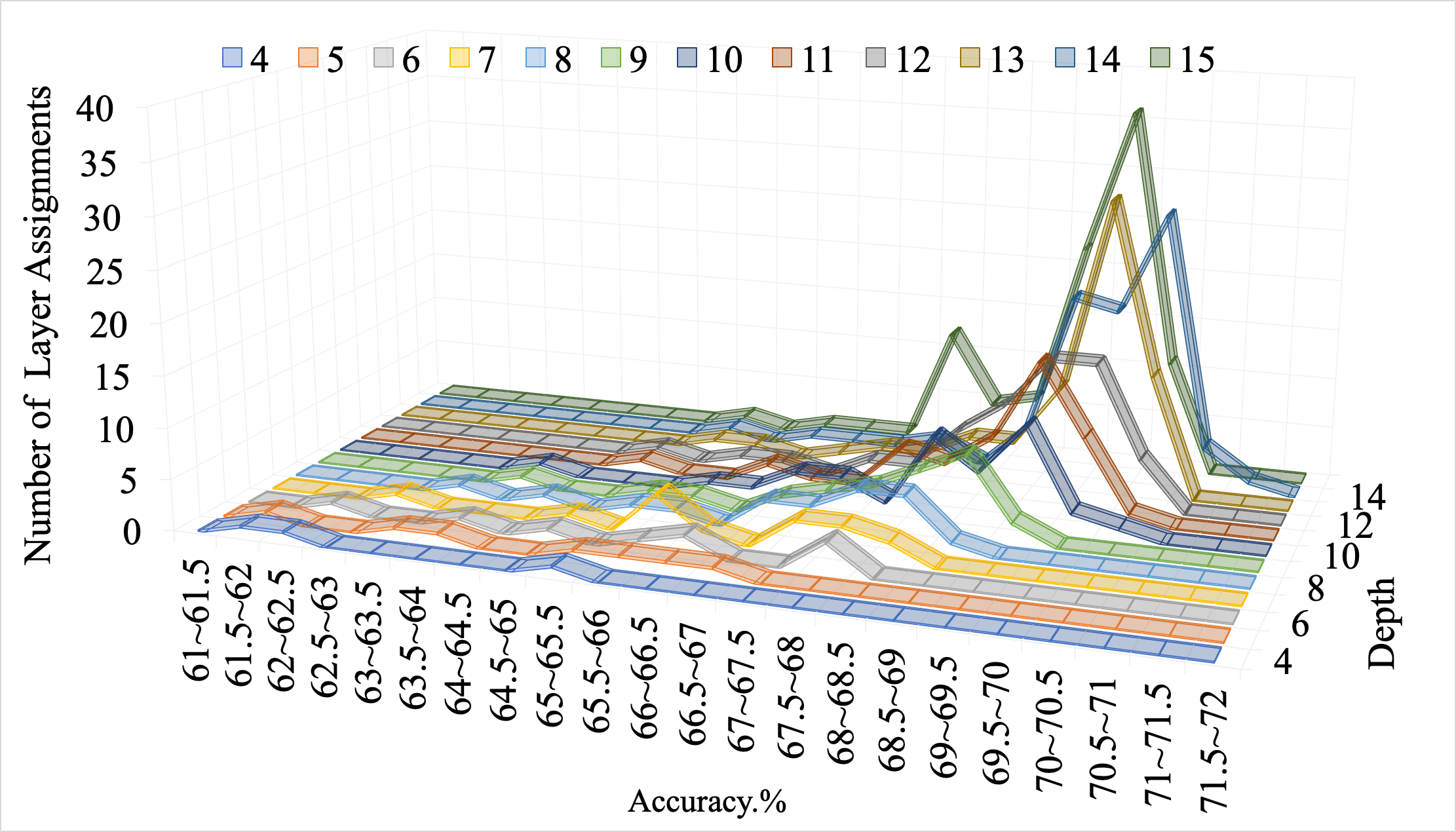}
\caption{Layer assignment distribution of residual networks on CIFAR-100. }
\label{Figure 5}
\end{figure}

\begin{figure}[!t]
\centering
\subfigure[]{
\label{fig:subfig:a} 
\includegraphics[width=.8\columnwidth]{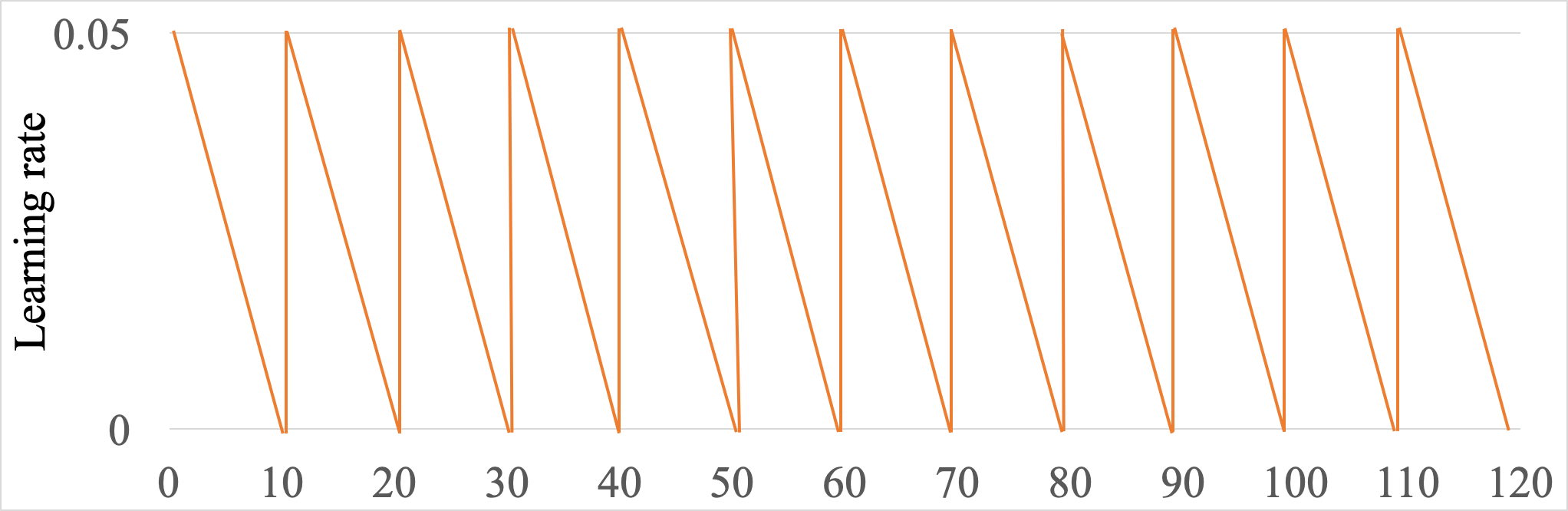}}
\subfigure[]{
\label{fig:subfig:b} 
\includegraphics[width=.8\columnwidth]{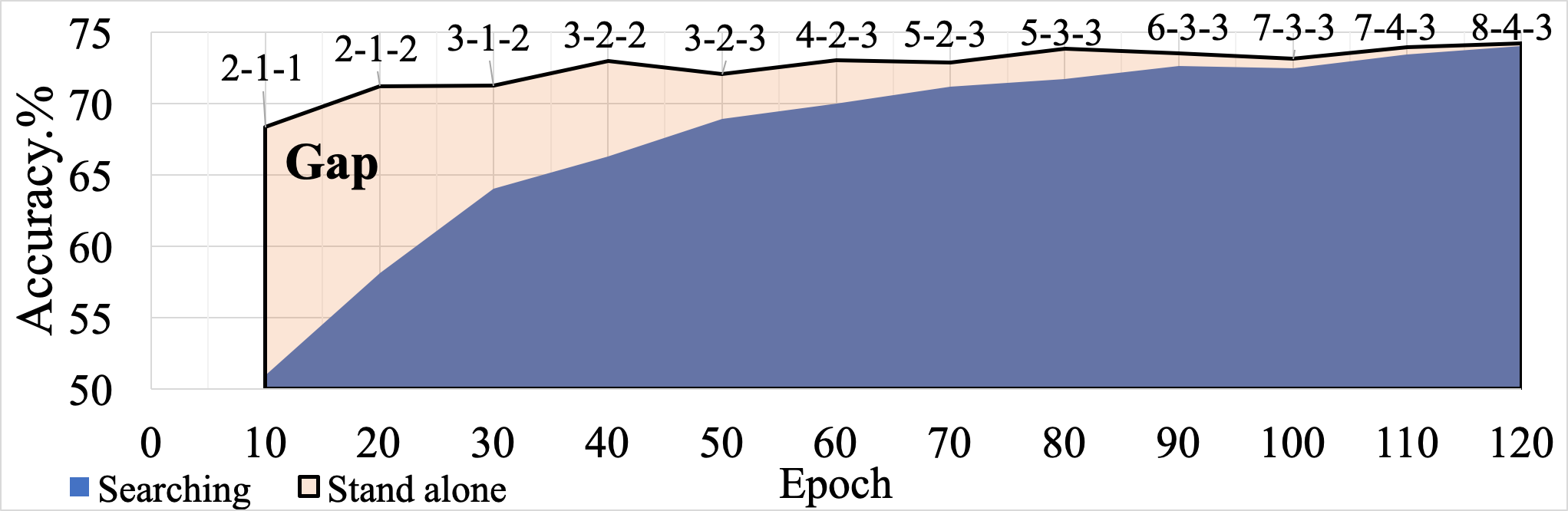}}
\subfigure[]{
\label{fig:subfig:b} 
\includegraphics[width=.8\columnwidth]{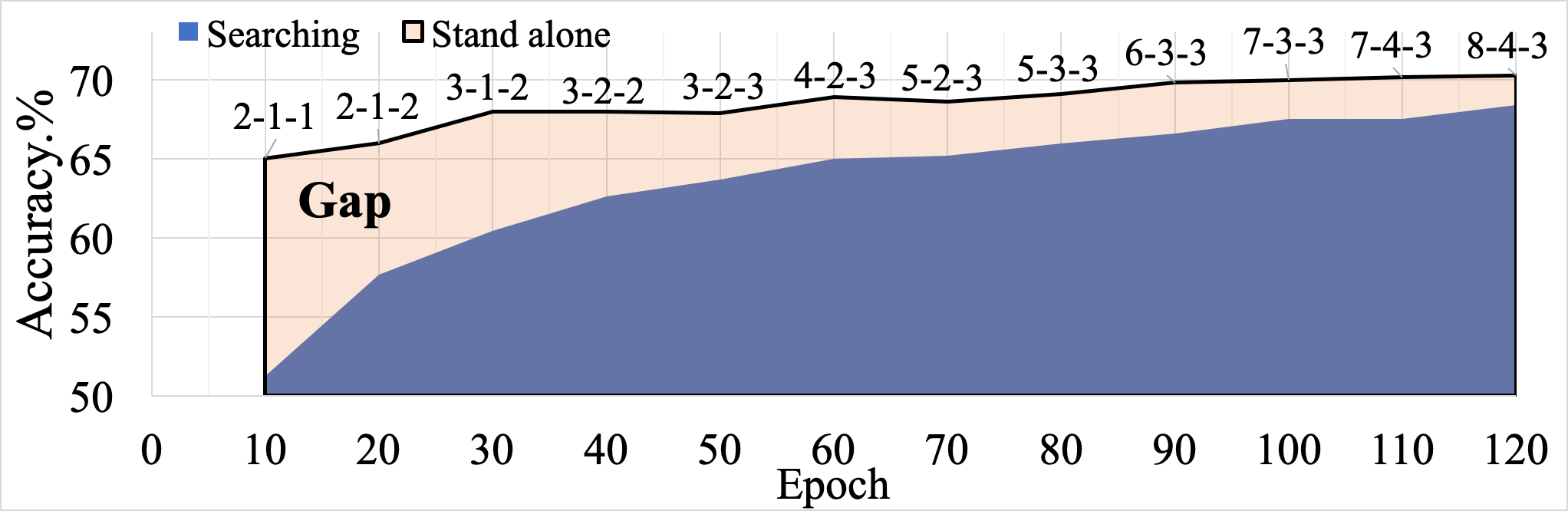}}
\caption{Searching procedure of plain network and residual network on CIFAR-100. (a) Learning rate varies with training epoch. (b) and (c) Searching accuracy of plain network and residual network vary with training epoch, and the accuracy of corresponding stand-alone networks. }
\label{Figure 6} 
\end{figure}

To provide arguments to Assumption 1, we build a layer assignments dataset on CIFAR-100. It consists of 908 different neural networks including both plain networks and residual networks, as well as their corresponding accuracy on the validation dataset. In this architecture dataset, we enumerate all the layer assignments for the network with $4$ to $15$ layers (or bottlenecks). After building the architecture dataset, we analyze the property of layer assignment from the following aspects:
\begin{itemize}
\item Given a fixed network depth, different layer assignments, different performance. As shown in Fig.\ref{Figure 1}, we collect the accuracy change of different layer assignments from shallow to deeper networks. Apparently, the performance gap is significantly large. Take network depth 11 as an example, the accuracy in plain network can change from 68.11\% to 74.37\%, and the accuracy in residual network can change from 64.92\% to 70.29\%. 
\item Different network configurations, different optimal layer assignments. We take out the optimal layer assignments for both plain networks and residual networks. As shown in Tab.\ref{table1}, for plain networks, they tend to place the layers in the shallow groups, while for residual networks, they tend to place the layers in the higher groups. It means that we cannot summarize a general layer assignment design manner for different network configurations.
\item Layer assignment distribution in each network depth. We visualize the layer assignment distribution in Fig.\ref{Figure 5}. The majority lie in the not bad region, while the percentage of the optimal ones is quite small, which indicates the difficulty of layer assignment search. As layer increases, the accuracy gap between the majority part and the optimal part is getting small since the influence of layer assignment is weakened by increasing network depth. This also provides an indirect proof for Randomly Wired Neural Networks\cite{xie2019exploring} that even randomly wiring can generate a not-bad network.
\item Neural inheritance relation among the optimal layer assignments from shallow networks to deeper networks. To discover the hidden relation among the layer assignments with different network depth. We take out top4 optimal layer assignments from the architecture dataset and plot them in Fig.\ref{Figure 4}. We can see that, the optimal layer assignments in deeper networks always inherit the configuration from shallow networks and further develop the optimal layer assignment to their network depth. It means that we can fast discover the optimal one in an extremely small network space if given the optimal layer assignment in shallow network. This is the key discovery in this paper and inspires our proposed LAS method.
\end{itemize}

\subsubsection{Automatic Layer Assignment Search.} We iteratively search the optimal layer assignment from shallow networks to deeper networks according to Algorithm. 1, and the searching procedure is shown in Fig.\ref{Figure 6}. We take out the win-out layer assignments in different network depths and use the accuracy trained from scratch to compare with each corresponding best layer assignment. As shown in Tab.\ref{table1}, the accuracy gaps between our searched networks and the best ones are quite small.

\subsection{Experiments on Tiny-ImageNet}
\subsubsection{Training Details.} We augment the training images by RandomResizedCrop function with default settings in Pytorch as well as randomly horizontal flipping. During searching process, we train the supernet with SGD optimizer with Nesterov momentum 0.9, batch size 256 and dropout ratio 0.5. The base learning rate is set to 0.05 with linear decaying policy in each step. Weight decay is set as 0.0001 for MobileNet-V1 and 0.0005 for ResNet-50. To compare the performance between our searched networks and the handcrafted ones more fairly after searching, we train both from scratch with exactly the same training configuration. It can decouple the influence of our special training manner used in searching process. Different from searching process, we train MobileNet-V1 and ResNet-50 for 200 epochs. The base learning rate is set to 0.1 and multiplied with a factor 0.1 at 100 epochs, 150 epochs and 175 epochs. 

\subsubsection{Searching Result on MobileNet-V1.} As shown in Tab.\ref{table2}, our searched layer assignment is [2, 5, 2, 3], which is totally different with handcrafted one [2, 2, 6, 2]. Ours surpasses it by a large margin with 1.42\% accuracy improvement.

\subsubsection{Searching Result on ResNet-50.} As shown in Tab.\ref{table2}, our searched one is [5, 6, 4, 1], and the handcrafted one is [3, 4, 6, 3]. Ours surpasses it by 0.4\% accuracy improvement. This improvement is less than MobileNet-V1, as we mention above that the impact of layer assignment is weakened by the improvement brought by the increasing network depth.

\subsection{Experiments on ImageNet}
\subsubsection{Training Details.} From the experiments on Tiny-ImageNet, we can see that the impact of layer assignment in shallow compact networks is more significant than deeper networks. Hence, we use MobileNet-V1 as a representative to further show the impact of layer assignment on the very large-scale dataset. We only use $1/6$ training datasets for searching. We use almost the same training configuration with the experiments on Tiny-ImageNet. Differently, weight decay here is set to 0.00004, the base learning rate is set to 0.6, and the batch size for every GPU is set to 128. We totally use 7 GPUs for training. 

\subsubsection{Searching Result on MobileNet-V1.} 
\begin{figure}[!t]
\centering
\includegraphics[width=.85\columnwidth]{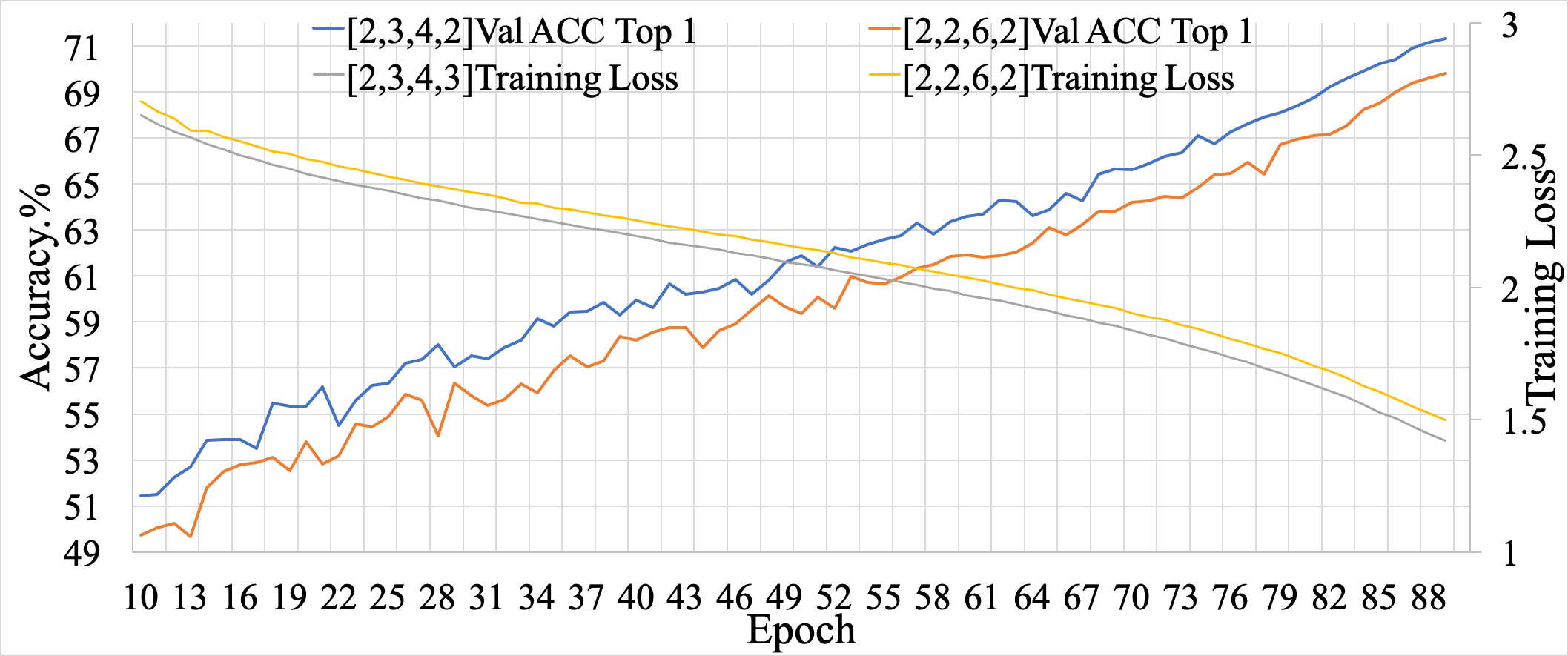}
\caption{The training procedures of our searched MobileNet-V1 and the handcrafted one.Both are trained from scratch with exactly the same training settings.}
\label{Figure 7}
\end{figure}

As shown in Tab.\ref{table3}, compared with handcrafted one [2, 2, 6, 2], our searched layer assignment only has a slight difference, which indicates that the designer for MobileNet-V1 may had already roughly search the layer assignment manually. Our searched result only moves two layers from 3th group to 2nd group and 4th group, respectively. Despite the slight difference, the performance of our searched one surpasses the handcrafted one by a large margin with an unexpected 1.5\% accuracy improvement! The detailed training procedures are shown in Fig.\ref{Figure 7}. Also, comparing the searched layer assignments for Tiny-ImageNet and ImageNet, the searching results for different datasets are totally different.

\subsubsection{Training Cost Analysis.} We do not use a lot of training time for searching since we only train $10$ epochs for each step, and the sub-networks in early-stages are shallow. As shown in Tab.\ref{table4}, our searching method only occupies half of training time compared with training a stand-alone network. 
\begin{table}[!t]
\caption{The performance comparison for the layer assignments searched for different network depth on CIFAR-100.}\smallskip
\centering
\resizebox{.95\columnwidth}{!}{
\smallskip\begin{tabular}{c|c|c|c|c|cc}
\hline
\multicolumn{1}{c|}{Network}&\multicolumn{2}{c|}{Search}&\multicolumn{2}{c|}{Best}&\multirow{2}{*}{\makecell[c]{Accuracy \\ Gap}}\\
\cline{2-5}
\multirow{15}{*}{plain}&\multicolumn{1}{c|}{\makecell[c]{Layer  Assignment}}&\multicolumn{1}{c|}{\makecell[c]{Accuracy}}&\multicolumn{1}{c|}{\makecell[c]{Layer Assignment}}&\multicolumn{1}{c|}{\makecell[c]{Accuracy}}&\\
\hline
&2-1-1&68.35&1-1-2&68.93&0.58&\\
&2-1-2&71.23&2-1-2&71.23&0&\\
&3-1-2&71.26&2-2-2&72.01&0.75&\\
&3-2-2&72.98&3-2-2&72.98&0&\\
&3-2-3&72.06&4-2-2&73.68&1.62&\\
&4-2-3&73.06&3-4-2&73.52&0.46&\\
&5-2-3&72.88&5-3-2&73.90&1.02&\\
&5-3-3&73.85&8-1-2&74.25&0.40&\\
&6-3-3&73.53&5-5-2&73.91&0.38&\\
&7-3-3&73.14&6-5-2&74.41&0.27&\\
&7-4-3&73.95&5-6-3&74.53&0.58&\\
&8-4-3&74.2&8-4-3&74.2&0\\
\cline{1-7}
\multirow{18}{*}{residual}&1-1-2&65.05&1-1-2&65.05&0\\
&1-1-3&66.97&1-1-3&66.97&0\\
&1-2-3&67.97&1-2-3&67.97&0\\
&1-2-4&67.91&1-1-5&68.30&0.39\\
&1-3-4&68.94&1-3-4&68.94&0\\
&2-3-4&68.64&1-3-5&69.23&0.59\\
&2-3-5&69.11&2-4-4&69.70&0.59\\
&2-4-5&69.83&3-3-5&70.29&0.46\\
&2-4-6&70.01&1-3-8&70.06&0.05\\
&2-5-6&70.34&3-5-5&70.43&0.09\\
&2-5-7&70.17&4-3-7&71.06&0.89\\
&2-5-8&70.28&5-4-6&71.06&0.78&\\
\hline
\end{tabular}
}
\label{table1}
\end{table}

\begin{table}[!t]
\caption{Searching results on Tiny-ImageNet}\smallskip
\centering
\resizebox{.95\columnwidth}{!}{
\smallskip\begin{tabular}{c|c|c|c|c}
\hline
\multicolumn{1}{c|}{Network}&\multicolumn{1}{c|}{Layer Assignment}&\multicolumn{1}{c|}{Accuracy(\%)}&\multicolumn{1}{c|}{FLOPs(M)}&\multicolumn{1}{c}{\#Params(M)}\\
\hline
MobileNet-V1(Our Impl.)&2-2-6-2&53.46&46.5&3.3\\
MobileNet-V1(Search)&\textbf{2-5-2-3}&\textbf{54.88}(1.42$\uparrow$)&46.7&3.5\\
\cline{1-5}
ResNet-50(Our Impl.)&3-4-6-3& 57.00&329.8&23.8\\
ResNet-50(Search)&\textbf{5-6-4-1}&\textbf{57.40}(0.4$\uparrow$)&329.8&13.4\\
\hline
\end{tabular}
}
\label{table2}
\end{table}

\begin{table}[!t]
\caption{MobileNet-V1 searching result on ImageNet}\smallskip
\centering
\resizebox{.95\columnwidth}{!}{
\smallskip\begin{tabular}{c|c|c|c|c}
\hline
Layer Assignment&Stand Alone Accuracy(\%)&Epoch&FLOPs(M)&\#Params(M)\\
\hline
2-2-6-2\cite{howard2017mobilenets}&70.6&$\gg$90&568.7&4.2\\
2-2-6-2(Our Impl.)&69.82&90&568.7&4.2\\
\textbf{2-3-4-3}(Search)&\textbf{71.32}&90&569.1&4.8\\
\hline
\end{tabular}}
\label{table3}
\end{table}

\begin{table}[!t]
\caption{The efficiency of MobileNet-V1 layer assignment search on ImageNet}\smallskip
\centering
\resizebox{.95\columnwidth}{!}{
\smallskip\begin{tabular}{c|c|c|c}
\hline
GPUs&\multicolumn{1}{c|}{Layer Assignment}&Search&\makecell[c]{Stand Alone}\\
\hline
\multirow{2}{*}{TITAN XP}&2-3-4-3&86.8 GPU-hours&173.9 GPU-hours\\
&2-2-6-2&-&183 GPU-hours\\
\hline
\end{tabular}
}
\label{table4}
\end{table}

\section{Conclusions}
In this paper, we discuss the impact of layer assignments to the network performance. We discover the neural inheritance relation among the networks with different layer assignments, which inspires us to propose an efficient one-shot layer assignment search approach via inherited sampling. We believe this intrinsic property can be extended to other dimensions of neural architecture and provides more insights to the NAS community.

\bibliographystyle{aaai} 
\bibliography{AAAI-MengR.7092}
\end{document}